\documentclass{article}
\usepackage{spconf,amsmath,graphicx}

\usepackage{amsfonts}
\usepackage{pifont}
\usepackage{bm}
\usepackage{comment}
\usepackage{xcolor}
\usepackage{url}

\newcommand{\defeq}{\stackrel{\mathrm{def}}{=}}

\newcommand{\E}[2]{\left\langle#2\right\rangle_{#1}}

\renewcommand{\vec}[1]{\bm{#1}}

\newcommand{\mat}[1]{\bm{\mathrm{#1}}}

\newcommand{\T}{\mathsf{T}}
\newcommand{\nm}[1]{{\mathtt{#1}}}

\newcommand{\ind}[1]{\mathbf{1}\left[ #1 \right]}
\newcommand{\nnet}[1]{\mathcal{#1}}
\newcommand{\ember}{\nnet{M}}
\newcommand{\vocab}{V}

\title{
KNOWLEDGE TRANSFER FROM LARGE-SCALE PRETRAINED LANGUAGE MODELS TO END-TO-END SPEECH RECOGNIZERS
}

\name{Yotaro Kubo, Shigeki Karita, Michiel Bacchiani}
\address{Google\\
    Shibuya Stream, 3--21--3 Shibuya, Shibuya-ku, Tokyo.}

\begin{document}
\ninept
\maketitle
\begin{abstract}
End-to-end speech recognition is a promising technology for enabling compact automatic speech recognition (ASR) systems since it can unify the acoustic and language model into a single neural network.
However, as a drawback, training of end-to-end speech recognizers always requires transcribed utterances.
Since end-to-end models are also known to be severely data hungry, this constraint is crucial especially because obtaining transcribed utterances is costly and can possibly be impractical or impossible.
This paper proposes a method for alleviating this issue by transferring knowledge from a language model neural network that can be pretrained with text-only data.
Specifically, this paper attempts to transfer semantic knowledge acquired in embedding vectors of large-scale language models.
Since embedding vectors can be assumed as implicit representations of linguistic information such as part-of-speech, intent, and so on, those are also expected to be useful modeling cues for ASR decoders.
This paper extends two types of ASR decoders, attention-based decoders and  neural transducers, by modifying training loss functions to include embedding prediction terms.
The proposed systems were shown to be effective for error rate reduction without incurring extra computational costs in the decoding phase.
\end{abstract}
\section{Introduction}
\label{sec:intro}

End-to-end techniques are known to be beneficial for realizing compact models for automatic speech recognition (ASR).
Since the compact models are advantageous both in terms of interface responsiveness and energy consumption, an accuracy improvement in end-to-end ASR is a crucial step for expanding the availability of speech recognition technologies in, for example mobile devices \cite{he2019streaming}.

One of the most promising approaches for improving end-to-end ASR accuracy is introducing additional training data since end-to-end models typically require large amount of training data for mitigating over-fitting.
However, obtaining labeled training data is often costly compared to obtaining unpaired audio-only and/or text-only data.
End-to-end systems model speech recognition as one unified model as opposed to a conventional system that factors the problem into several models and separates the language model (LM) component from the acoustic model (AM) component.
Hence, training from text-only data can only be realized in the form of an external LM which leads to additional complexity and run-time costs.

Recently, a series of important methods for utilizing unpaired audio-only data is proposed as pretraining methods of the end-to-end models \cite{oord2018representation,baevski2020wav2vec,zhang2020pushing}.
Those methods pretrain the encoder part of end-to-end ASR models by introducing a surrogate objective function based on mutual information between instantaneous feature representation and global context representation.
Since those pretraining processes can be performed with audio-only data, those methods are promising for relaxing the limitation due to availability of paired training data.

Previous work has shown an effective way to make use of unpaired audio data, the work here focus on a way to make use of unpaired text data without raising the model size and inference complexity.
There are several methods for integrating external language models trained on text-only data into end-to-end speech recognition \cite{toshniwal2018comparison}.
However, since language models are usually very large in terms of model size and computational complexity, the use of external LMs makes the model less compact and raises the computation cost.
Recent language models, such as \textit{bidirectional encoder representations from transformers} (BERT) \cite{devlin2019bert}, involve typically several hundred millions of parameters, and integrating it is computationally prohibitive for small devices even though the language model performs well and is in that respect desirable.

In this paper, learning from text-only data is achieved by designing multi-task learning that performs knowledge distillation \cite{hinton2015distilling}.
Applications of transfer learning do not require any additional computational cost in the recognition phase, and thus it is advantageous for models being deployed to small devices.
There are several recent attempts to distill knowledge from external models into end-to-end speech recognizers \cite{kurata2020knowledge,panchapagesan2021efficient,futami2020distilling,bai2021fast}.
The prior studies that are most related to our approach are based on distillation from a masked language model \cite{futami2020distilling,bai2021fast}.
For transferring knowledge from a pretrained BERT model, \cite{futami2020distilling} minimizes the KL-divergence between the token posterior distribution from the BERT masked language model and the ASR model.
In contrast, our approach focuses on using embedding vectors rather than posterior distribution of the masked LM.
As suggested in the experiments in the original BERT paper \cite{devlin2019bert}, embedding vectors are believed to be a compact representation containing richer information compared to the posterior distributions of tokens.
On the other hand, knowledge transfer from embedding vectors is shown to be effective for improving performance of non-auto-regressive speech recognizers \cite{bai2021fast}.
In this paper, the effectiveness of knowledge transfer from embedding vectors is verified with attention- \cite{chan2016listen} and transducer-based \cite{graves2012sequence} speech recognition models.
Since the transducer-based approach is popular in ASR for computationally limited devices as described in \cite{he2019streaming}, it is important to analyze the effectiveness of knowledge transfer for such models.

\section{Knowledge distillation from embeddings} \label{sec:method}

This section describes methods to enhance existing end-to-end ASR neural networks to multitask models that predict precomputed token-embedding vectors in addition to the token distributions.
In this section, each element in the training dataset is denoted as a triple $(\mat{X}, \vec{y}, \mat{E})$  of the feature vector sequence  $\mat{X}$, token sequence $\vec{y} = (y_1, y_2, \cdots)$, and the sequence of token embedding vectors $\mat{E} = (\vec{e}_1, \vec{e}_2, \cdots)$.
Each token $y_i \in \vocab$ represents a subword token, typically a word-piece \cite{schuster2012japanese}, of the transcription.
The embedding vectors $\mat{E}$ and the token sequences $\vec{y}$ are related by a word embedding function $\ember: \vocab^{N} \to ( \mathbb{R}^{D^{\nm{Emb}}} )^{N}$ as $\mat{E} = \ember(\vec{y})$, and therefore the numbers of word tokens and the token-embedding vectors in the same triple are the same.
In this paper, $\ember$ is represented in a pretrained neural network, and its parameters are not fine-tuned.
Specifically, in the experimental section below, we used BERT embeddings as $\ember$.

\subsection{Attention-based decoder with auxiliary regression}

\begin{figure*}[bt]
    \centering
    \includegraphics[width=0.9\textwidth]{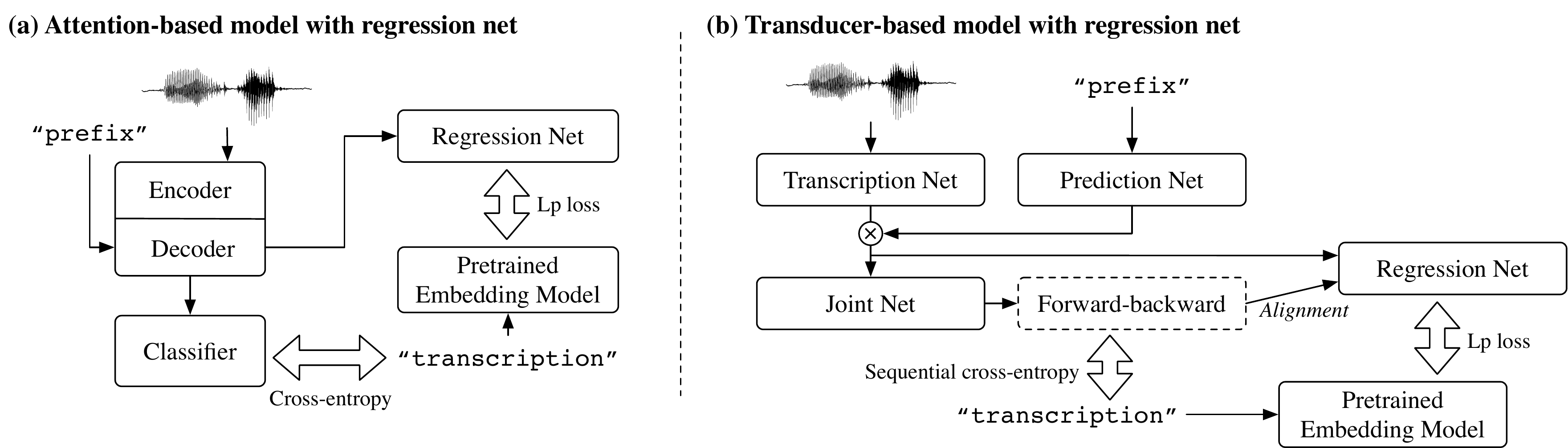}
    \caption{Block diagrams of (a) attention-based model with a regression net, and (b) transducer-based model with a regression net. } \vspace{-2mm}
    \label{fig:diagram}
\end{figure*}

An attention-based auto-regressive decoder predicts the probability distribution of the next token $y_i$ using the current decoder state $\vec{\phi}(\mat{X}, \vec{y}_{1:i-1})$ as,
\begin{equation}
    p(y_i \mid \mat{X}, \vec{y}_{1:i-1}) \propto \exp \left( \vec{\lambda}_{y_i}^{\T} \vec{\phi}(\mat{X}, \vec{y}_{1:i-1}) \right).
\end{equation}
Here, $\vec{\lambda}_{y_i}$ is the parameter vector used for computing a logit for $y_i$,
and $\vec{\phi}(\mat{X}, \vec{y}_{1:i-1}) \in \mathbb{R}^{D}$ is a vector referred to as the ``pre-softmax activation''.
Typically, $\vec{\phi}$ is equipped with an attention mechanism to summarize the encoded representation of $\mat{X}$.

A cross-entropy loss is defined by using the above prediction as,
\begin{equation}
    L^{\nm{XEnt}}(\mat{X}, \vec{y}) = - \sum_{i} \log p(y_i \mid \mat{X}, \vec{y}_{1:i-1}).
\end{equation}
The conventional training for end-to-end models with an attention-based decoder minimizes the expectation of this objective function using a gradient-based method such as Adam \cite{kingma2014adam}.

In this paper, aiming at transferring knowledge from token-embedding vectors to the decoder state vector $\vec{\phi}$, a regression neural network $\nnet{R}: \mathbb{R}^{D} \to \mathbb{R}^{D^{\nm{Emb}}}$ is introduced.
With this additional component, the following loss function $L^{\nm{Emb}}$ based on the distances between the estimated embeddings and the precomputed embeddings $\vec{e}_i$ is introduced as an auxiliary loss function to be minimized.
\begin{equation}
    L^{\nm{Emb}}(\mat{X}, \vec{y}, \mat{E}) = \sum_{i}
    d(\nnet{R} ( \vec{\phi}(\mat{X}, \vec{y}_{1:i-1}) ), \vec{e}_i),
\end{equation}
where $d$ is a differentiable distance function such as Lp distances.

With the above auxiliary loss function, the loss function to be minimized in training can be defined as,
\begin{equation}
    L^{\nm{XEnt}}(\mat{X}, \vec{y}) 
    + \sigma L^{\nm{Emb}}(\mat{X}, \vec{y}, \mat{E}).
    \label{Eq:MultitaskLoss}
\end{equation}
Here, $\sigma$ is a hyper-parameter that controls the relative importance of the auxiliary task.
Setting $\sigma = 0$ reduces the training process to the conventional cross-entropy on the token probabilities alone.

Fig \ref{fig:diagram}-(a) depicts neural net architecture for implementing training with this loss function.
The ``Regression Net'' and ``Pretrained Embedding Model'' components in the figure can be pruned in the recognition phase.
Therefore, the total computational complexity at the recognition phase is not increased by this modification.

\subsection{Transducer-based decoder with auxiliary regression}
\label{sec:method:transducer}

Neural transducers (also known as recurrent neural network transducers (RNN-Ts)) \cite{graves2012sequence} model a probability distribution of alignment variables $\vec{z} = (z_1, z_2, \cdots) \in (\vocab \cup \left\{ \varphi \right\})^{T + N}$ where $T$ and $N$ are the lengths of a feature vector and a word sequence, respectively, and $\varphi$ is a blank symbol.
The token sequence $\vec{y}$ is considered to be extracted by using a \textit{blank removal function} $\nnet{B}$ that removes $\varphi$ from the alignment sequence $\vec{z}$.
Therefore, the predictive distribution of the token sequences can be expressed by marginalizing the alignment variables, as follows:
\begin{equation}
    p(\vec{y} \mid \mat{X}) = \sum_{\vec{z}'} \ind{\nnet{B}(\vec{z}') = \vec{y}} p(\vec{z}' \mid \mat{X}). \label{Eq:MarginalizeAlignment}
\end{equation}
Here, $\ind{P}$ is an indicator function whose value is $1$ if the predicate $P$ is true, and $0$ otherwise.
It should be noted that $\ind{\nnet{B}(\vec{z}') = \vec{y}}$ can be viewed as a hard-coded model of $p(\vec{y} \mid \vec{z}')$.

The probability distribution over the alignment $p(\vec{z} \mid \mat{X})$ is computed depending on outputs of two neural networks, the transcription and prediction networks.
The transcription network computes feature vectors $\vec{\phi}_t$ for each frame $t$.
The prediction network computes the representation from the prefix of a (hypothesis) token sequence.
In the training phase with teacher forcing, the prediction network computes output vectors $\vec{\psi}_{i}$ for each token  $y_i$ using the token prefix $\vec{y}_{1:i-1}$.
In this paper, $\vec{\phi}_t$ and $\vec{\psi}_{i}$ are called acoustic and language feature vectors.

Using those two feature vectors, the joint network $\mathcal{J}$ computes the probability distribution over the alignment variable $\vec{z}$ as,
\begin{equation}
    p(z_{n} = k \mid \vec{z}_{1:n-1}, \mat{X}, \vec{y}) = \nnet{J}_k(\vec{\phi}_{1 + \tau(\vec{z}_{1:n-1})}, \vec{\psi}_{1 + \iota(\vec{z}_{1:n-1})}),
    \label{Eq:RNNTAlignmentEmission}
\end{equation}
where $\nnet{J}_k(.)$ is the $k$-th element of the output vector of the joint network, $\tau(\vec{z}_{1:n-1})$ and $\iota(\vec{z}_{1:n-1})$ are the numbers of blank $\varphi$ and non-blank symbols, respectively, in the prefix $\vec{z}_{1:n-1}$.

\subsubsection*{Joint regression network}

Here, similar to attention-based decoder modeling, a regression network $\nnet{R}$ is introduced to predict token-embedding vectors $\mat{E}$.
The auxiliary loss can be defined as an expectation of distances as,
\begin{equation}
    L^{\nm{Emb}}(\mat{X}, \vec{y}, \mat{E}) = \sum_{i}
    \E{q_i(t \mid \mat{X}, \vec{y})}{
    d(
    \nnet{R} ( \vec{\phi}_t, \vec{\psi}_{i}) ,
    \vec{e}_i
    )}. \label{Eq:TransducerAuxLoss}
\end{equation}
It should be noted that the regression network here takes two input vectors similar to the joint network.
The alignment probability $q_i(t \mid \mat{X}, \vec{y})$ is a probability of $t$-th acoustic feature vector being consumed after processing the prefix $\vec{y}_{1:i-1}$, and can be expressed as,
\begin{equation}
\begin{aligned}
&    q_i(t \mid \mat{X}, \vec{y}) \defeq \\
&    \sum_{\vec{z}'} \sum_{n} \ind{\iota(\vec{z}'_{1:n}) = i - 1 \land\tau(\vec{z}'_{1:n}) = t - 1} p(\vec{z}' \mid \mat{X}, \vec{y}).
\end{aligned}
\end{equation}
The expectation over this posterior distribution can be computed by reusing the results of the forward-backward algorithm used in the computation of the conventional transducer loss function.
It should be noted that gradient information was not propagated through this $q$ function in the experimental section below, to ensure that auxiliary loss does not affect to alignment computation.

Fig \ref{fig:diagram}-(b) depicts an example neural net architecture for this training.
The expectation in Eq. \eqref{Eq:TransducerAuxLoss} can efficiently be computed by reusing the alignment variable (specifically, forward and backward score) from the forward-backward computation in the main transducer loss.

\subsubsection*{Token-synchronous regression network}

As an alternative to the loss function described above, we investigated another loss that has a single regression result per token.

This variant approximates the expectation of the loss functions in Eq. \eqref{Eq:TransducerAuxLoss} by using the expectation of the acoustic features as,
\begin{equation}
    L^{\nm{Emb}}(\mat{X}, \vec{y}, \mat{E}) \simeq \sum_{i}
    d(
    \mathcal{R} ( 
    \E{q_i(t \mid \mat{X}, \vec{y})}{
    \vec{\phi}_t}, \vec{\psi}_{i}) ,
    \vec{e}_i
    ). \label{Eq:TransducerAuxLossSync}
\end{equation}
This approximated loss is equivalent to Eq. \eqref{Eq:TransducerAuxLoss} when an L2 distance metric and a linear network are used as $d$ and $\mathcal{R}$, respectively.

This approximation reduces the training memory consumption regarding the regression network from $O(TN)$ to $O(N)$.
Furthermore, the regression network in this architecture is trained to make a single prediction per token whereas the joint regression network in Eq. \eqref{Eq:TransducerAuxLoss} makes multiple predictions minimizing the expectation of the loss.
Therefore, this approximation is also beneficial as the token-wise predictions can be helpful for subsequent processing of the ASR results.

\section{Experiments}

We verified the effectiveness of the proposed methods by training models on the LibriSpeech dataset \cite{panayotov2015librispeech} (960h of transcribed speech), and using a BERT \cite{devlin2019bert} language model pretrained with the BooksCorpus (800M words) \cite{zhu2015aligning} and Wikipedia text data (2500M words).

\subsection{Experimental Setup}

The attention-based model we used was based on bi-directional long short-term memory (BLSTMs).
The configurations for the BLSTM encoder and decoder were set to be identical with those of \cite{irie2019choice}.
The transducer-based model employed a Conformer encoder \cite{gulati2020conformer}.
The encoder consisted of 17 Conformer blocks, and each block was equipped with an 8-head dot-attention, 512-channel 1d-convolution, and 2048-dim feed-forward module.
The prediction network in the transducer-based model was an LSTM consisting of a 640-dimensional (uni-directional) long short-term memory (LSTM) layer and a 128-dimensional token-embedding layer.
The joint network in the transducer-based model was a feed-forward network with a 640-dim hidden layer with $\tanh$-activation.

SpecAugment \cite{park2019specaugment} was applied in the transducer-based systems.
The numbers of time- and frequency-masks were set to 10 and 2, respectively.
The lengths of the time- and frequency-masks were set to $0.05T$ and $27$, respectively, where $T$ is the length of the input feature vector sequence.
Both for the attention and transducer-based models, Adam \cite{kingma2014adam} was used as the optimization method.
The batch sizes were set to $512$ and $2048$ for the attention and transducer-based models, respectively.
The optimization results of transducer-based systems was smoothed by computing  exponential moving average (EMA) of the parameter trajectory.
The decay rate for EMA was set to $0.9999$.
For each training configuration, the best checkpoint was chosen based on the WERs observed on the \texttt{devother} partition.

The weight for the auxiliary task ($\sigma$ in Eq. \eqref{Eq:MultitaskLoss}) was varied for analyzing the effect of the auxiliary task.
For attention-based models, the weights were selected from  $\sigma \in \{2^{0}, 2^{-1}, 2^{-2}, 2^{-3}, 2^{-4}\}$, and for transducer-based models,
the weights were selected from $\sigma \in \{10^{-1}, 10^{-2}, 10^{-3}, 10^{-4}\}$.
As it is mentioned above, setting $\sigma=0$ is equivalent with omitting the auxiliary task, and was considered as a baseline setting.

The distance metric used in the auxiliary loss function was chosen as the L1 distance between the predicted and target vectors normalized (divided) by the dimensionality of the embedding vectors ($D^{\nm{Emb}} = 768$).
The regression network $\nnet{R}$ was defined by a single affine transformation.
For the transducer-based decoder, we first used a token-synchronous regression module, i.e. the loss function in Eq. \eqref{Eq:TransducerAuxLossSync}.
As described in the previous section, Eqs. \eqref{Eq:TransducerAuxLossSync}  and \eqref{Eq:TransducerAuxLoss} are identical if we use an L2 distance and a linear regression network.
We used an L1 distance in this experiment and expected a subtle difference which we analyzed further in the results below.

The pretrained BERT model we used is identical with one available on TFHub \cite{tfhub_bert}.
This BERT embedding module consists of 12 transformer blocks with 768 hidden units and 12 self-attention heads.
For generating the target embedding vector sequences for training, the transcriptions in the training data were tokenized and fed into the BERT embedding module.
In this experiment, since we adopted BERT tokens using 30k word-piece vocabulary, the ASR models were also trained with this tokenization method.
Since word-piece models used in the baseline ASR model did not match those used for BERT modeling, we also investigated the effect of using word-piece models developed for BERT in ASR \footnote{We kept special tokens for BERT modeling, such as \texttt{[CLS]} and \texttt{[SEP]} as ASR output target sequence.}.

\subsection{Main results}

Table \ref{tab:librispeech_wer_las} shows the word error rates (WERs) of the baseline and the proposed training methods applied to attention-based models.
By comparing ``Baseline'' and ``BERT tok.'', we observe that the use of a different tokenization strategy did not degrade the word error rates significantly.
By increasing, the auxiliary task weight $\sigma$, we observed a word error rate reduction in all the test sets.
The best configuration for all the test sets is obtained when $\sigma$ was set to $2^{-2}$ where the relative error rate reduction was -16.7\% and -10.6\% for the \texttt{testclean} and \texttt{testother} datasets.
By increasing the auxiliary weight further, we observed a performance degradation.

\begin{table}[tb]
    \centering
    \begin{tabular}{l|r|rrrr}
        \hline \hline
        LAS & & \multicolumn{2}{c}{dev} & \multicolumn{2}{c}{test} \\
        & $\sigma$ & {clean} & {other} & {clean} & {other} \\
        \hline
        Baseline & -- & 4.5 & 13.5 & 4.7 & 13.7 \\
        BERT tok. & 0 & 4.6 & 13.2 & 4.8 & 14.1 \\
        \hline
        +BERT reg. & 0.0625 & 4.2 & 12.1 & 4.3 & 13.2 \\
        & 0.125 & 4.0 & 11.8  & 4.2 & 12.7 \\
        & 0.25 & {\bf 3.9} & {\bf 11.6} & {\bf 4.0} & {\bf 12.6} \\
        & 0.5 & 4.1 & 12.3 & 4.2 & 13.2\\
        & 1.0 & 4.2 & 13.0 & 4.2 & 13.8 \\
        \hline
    \end{tabular}
    \caption{WERs of attention-based models: "BERT tok." denotes a system with the baseline training method and word-piece model from pretrained BERT model. "+BERT reg." denotes the systems with auxiliary regression task with varying task weights.}
    \label{tab:librispeech_wer_las}
\end{table}

Table \ref{tab:librispeech_wer_confo} shows the word error rates of the baseline and the proposed training methods applied to transducer-based models.
The best configuration for \texttt{devclean} is obtained when $\sigma$ was set to $10^{-2}$ where the relative error rate reduction was -3.7\% and -11.3\% for \texttt{testclean} and \texttt{testother} datasets.

\begin{table}[tb]
    \centering
    \begin{tabular}{l|r|rrrr}
        \hline \hline
        ConformerS & & \multicolumn{2}{c}{dev} & \multicolumn{2}{c}{test} \\
        & $\sigma$ & {clean} & {other} & {clean} & {other} \\
        \hline
        Baseline & -- & 2.6 & 6.8 & 2.7 & 6.3 \\
        BERT tok. & 0 & 2.6 & 5.7 & 2.7 & 6.2 \\
        \hline
        +BERT reg. & 0.0001 & 2.5 & 5.5 & 2.6 & 5.6 \\
        & 0.001 & 2.4 & {\bf 5.2} & 2.7 & 5.7 \\
        & 0.01 & {\bf 2.4} & 5.3 & {\bf 2.6} & {\bf 5.5} \\
        & 0.1 & 2.4 & 5.3 & 2.8 & 5.9 \\
        \hline
    \end{tabular}
    \caption{WERs of transducers (see also Table \ref{tab:librispeech_wer_las} for legends.)}
    \label{tab:librispeech_wer_confo}
\end{table}

From both applications, we confirmed that the multitask learning with an auxiliary regression to the word embeddings were beneficial for transferring semantic knowledge from the pretrained BERT language model.
Since setting the auxiliary weight $\sigma$ to zero converges to the single task model, we observed WER convergence to the baseline model by reducing $\sigma$.
We also observed subtle differences due to the tokenization method we employed.
The WordPiece models we took from the BERT model performed better with our Conformer models. This might be due to the fact that the Conformer has more expressive ability and the larger inventory of tokens in BERT-based tokenization worked well with this setting.

\subsection{Comparison between token-synchronous regression and joint regression net}

In the previous section, we verified the effectiveness of our method applied to transducer-based decoders using token-synchronized regression shown in Eq. \eqref{Eq:TransducerAuxLossSync}.
Even though the difference between the two variants proposed in Section \ref{sec:method:transducer} is subtle, they are not identical in this experiment since an L1 distance is used as the distance metric.

\begin{table}[tb]
    \centering
    \begin{tabular}{l|r|rrrr}
        \hline \hline
        ConformerS & & \multicolumn{2}{c}{dev} & \multicolumn{2}{c}{test} \\
        & Best $\sigma$ & {clean} & {other} & {clean} & {other} \\
        \hline
        BERT tok. & 0 & 2.6 & 5.7 & 2.7 & 6.2 \\
        \hline
        Token-sync. regr. & 0.01 & 2.4 & 5.3 & 2.6 & 5.5 \\
        Joint regr. & 0.01 & 2.5 & 5.1 & 2.7 & 5.8 \\
        \hline
    \end{tabular} 
    \caption{Comparison between token-synchronized regression and joint regression networks.} \label{tab:joint_vs_aligned}
\end{table}

Table \ref{tab:joint_vs_aligned} summarizes the results of the model with joint regression and with token-synchronized regression networks.
The best auxiliary weight was chosen based on the results on \texttt{devother}.
From the table, we could not observe clear performance difference between the loss functions compared (Eqs. \eqref{Eq:TransducerAuxLoss} and \eqref{Eq:TransducerAuxLossSync}).

As mentioned in Section \ref{sec:method:transducer}, the token-synchronized regression network has qualitative advantages.
Therefore, in this setting, we confirmed that the use of token-synchronized regression network was more relevant.
However, the accuracy difference might be larger if we were to use more complicated regression network than a linear regression network.

\subsection{Results with pretrained acoustic encoders}

As an additional experiment, we combined the proposed training method with a large-scale pretrained acoustic encoder.
Application for a model like that is most promising as it opens the door for end-to-end models to use both unpaired audio and text data.
Specifically, we evaluated the proposed methods with Conformer-based wav2vec-2.0 pretraining method \cite{zhang2020pushing}.
The pretrained encoder and the decoder setting were identical with those of ``Conformer-XL'' in \cite{zhang2020pushing} pretrained with LibriLight dataset (60k hours).
As in the previous experiments, the hyper-parameters were chosen to minimize the word-error rates over the \texttt{devother} test set.

\begin{table}[tb]
    \centering
    \begin{tabular}{l|r|rrrr}
        \hline \hline
        Pretrained-ConfXL & & \multicolumn{2}{c}{dev} & \multicolumn{2}{c}{test} \\
        & Best $\sigma$ & {clean} & {other} & {clean} & {other} \\
        \hline
        Baseline & 0 & 1.77 & 3.47 & 1.75 & 3.56 \\
        +BERT reg. & 0.01 & 1.73 & 3.40 & 1.80 & 3.51 \\
        \hline
    \end{tabular}
    \caption{WERs with pretrained acoustic encoders}
    \label{tab:librispeech_wer_w2v2}
\end{table}

Table \ref{tab:librispeech_wer_w2v2} shows the WERs of the proposed training methods combined with the pretrained encoder.
We confirmed that the proposed method could further reduce the word error rate even from the strong baseline with a pretrained encoders.
However, the advantage is relatively small and this suggest that a part of semantic information might already be captured in wav2vec-2.0 training even without using a transcription.
Future work will focus on improving the configuration for this type of combination.

\section{Conclusion} \label{sec:conclusion}

This paper proposed a simple method to utilize text-only data and pretrained word embedding models for enhancing performance of speech recognition without increasing the model size.
The proposed method is based on multi-task learning consisting of a main speech recognition task and an auxiliary word-embedding regression task.
The multi-task learning method is applied to both attention and transducer-based end-to-end speech recognizers.
We confirmed that, with including the auxiliary task as our methods propose, we were able to improve the transcription accuracies of the resulting systems.
For attention-based models, the error reductions were 16.7\% and 10.6\% for the \texttt{testclean} and \texttt{testother} datasets, respectively.
For transducer-based models, the error reductions were 3.7\% and 11.3\% for the \texttt{testclean} and \texttt{testother} datasets, respectively.

Future work will focus on the usability of token-embeddings, obtained as a by-product of this model, for down-stream natural language processing tasks.
Even though we confirmed that a training process with the proposed method can reduce the distance metric between the estimated and precomputed token-embedding vectors, it is not clear if the estimated vectors are beneficial for, for example, an utterance classification task.

\bibliographystyle{IEEEbib}
\bibliography{refs}
\end{document}